\documentclass{article}

\usepackage{arxiv}

\usepackage{graphics} 
\usepackage{graphicx}

\usepackage{epsfig} 
\usepackage[utf8]{inputenc} 
\usepackage[T1]{fontenc}    
\usepackage{hyperref}       
\usepackage{url}            
\usepackage{booktabs}       
\usepackage{amsfonts}       
\usepackage{nicefrac}       
\usepackage{microtype}      
\usepackage{lipsum}

\title{Towards a Science of Resilient Robotic Autonomy}

\author{
  Kostas Alexis\thanks{www.autonomousrobotslab.com} \\
  Department of Computer Science and Engineering\\
  Autonomous Robots Lab\\
  University of Nevada, Reno \\
  1664 N. Virginia St., Reno, NV, USA \\
  \texttt{kalexis@unr.edu} \\
}

\begin{document}
\maketitle

\begin{abstract}
  This discussion paper aims to support the argument process for the need to develop a comprehensive science of resilient robotic autonomy. Resilience and its key characteristics relating to robustness, redundancy, and resourcefulness are discussed, followed by a selected - but not exhaustive - list of research themes and domains that are crucial to facilitate resilient autonomy. Last but not least, an outline of possible directions of a new and enhanced design paradigm in robotics is presented. This manuscript is intentionally short and abstract. It serves to open the discussion and raise questions. The answers will necessarily be found in the actual process of conducting research by the community and in the framework of introducing robotics in an ever increasing set of real-life use cases. Its current form is based on thoughts identified within the ongoing experience of conducting research for robotic systems to gain autonomy in certain types of extreme environments such as subterranean settings, nuclear facilities, agriculture areas, and long-term off-road deployments. The very context of this document will be subject to change and it will be iteratively revisited. 
\end{abstract}


\section{Introduction}\label{sec:intro}

\begin{figure}[h!]
\centering
    \includegraphics[width=0.99\columnwidth]{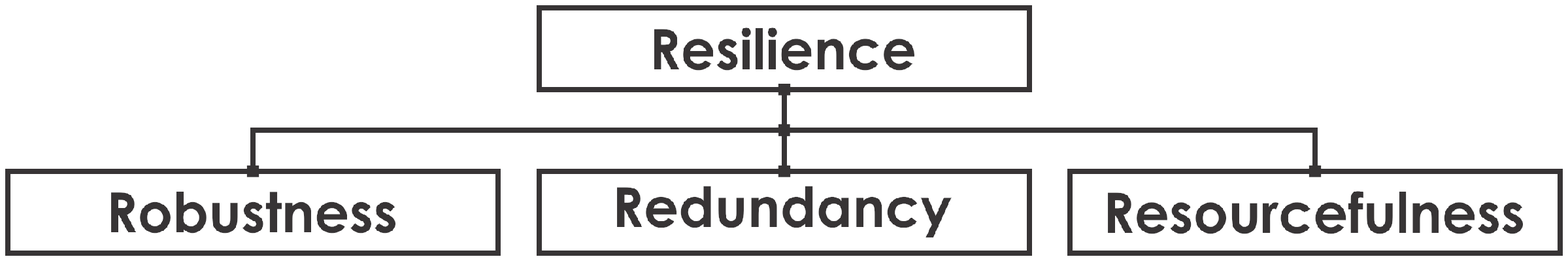}
\caption{The elements of resilience: robustness, redundancy, and resourcefulness. }\label{fig:elementsresilience}
\end{figure}

A system, individual or system-of-systems, presents resilience if it demonstrates the characteristics of a) Robustness, b) Redundancy, and c) Resourcefulness. This organization is inspired by risk analysis and mitigation studies~\cite{howell2013global} and is visually depicted in Figure~\ref{fig:elementsresilience}. Below we provide a brief notional definition of these key features and a broad outline of how this may relate to robotic functionality. An outline of how resilience relates to key operations and functional loops of a robotic system or system-of-systems follows. 

%
%

\textbf{Robustness and Robust Performance:} Robustness incorporates the concept of reliability and refers to the ability to absorb and withstand disturbances and crises in unpredictable situations. The assumptions underlying this component of resilience primarily relate to: 1) if fail-safes and firewalls are designed, and 2) if the decision-making and functioning of the system become more modular and individually more self-reliable then potential damage to one part of the system is less likely to spread far and wide leading to the ``collapse'' of the system’s functionality. 

\textbf{Redundancy:} Redundancy involves having excess capacity and back-up systems, which enable the maintenance of core functionality in the event of disturbances and sub-system failures. This component assumes that a system will be less likely to experience a collapse in the wake of stresses or failures of some of its subcomponents, if the design of that system’s architecture and modules incorporates a diversity of overlapping subsystems (e.g, sensors, actuators), methods, policies, strategies or last-resort techniques to accomplish objectives and fulfill mission purposes. 

\textbf{Resourcefulness:} Resourcefulness means the ability to adapt to changes, uncertainties and crises, respond flexibly and - when possible - transform a negative impact into a positive. For a system to be adaptive means that it has inherent flexibility, which is crucial into enabling the ability to inject and embed system resilience. The assumption underlying this component of resilience is that if systems can utilize and assume trust on not only a single solution - a monolithic architecture for their operation - but a set of different combinations and utilizations of their subsystems (e.g., fusion of a subset of their sensing modalities) then they are more likely to ensure reliable operation on the basis of new, novel and diverse solutions that exploit in different ways their elementary capacities (e.g., in sensing, actuation). This in turn means greatly reduced likelihood of a full and complete system failure. Self-organization, adaptive use of onboard resources and varying architecture techniques are all instances of resourcefulness. 

\begin{figure}[h!]
\centering
    \includegraphics[width=0.99\columnwidth]{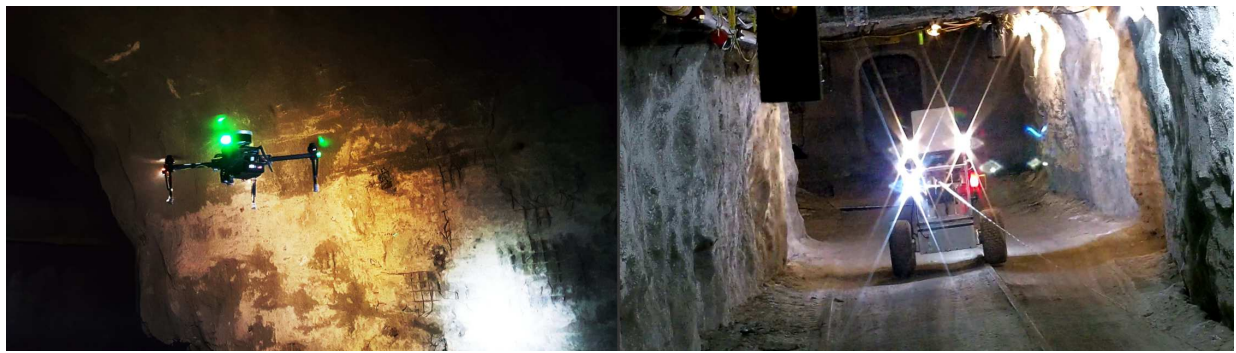}
\caption{Two robotic deployments inside underground mines. Missions of that type challenge the resilience of robotic systems. A Science of Resilient Robotic Autonomy is needed to address how to think and design for robotic systems, systems-of-systems and ``mosaics'' able to adjust to any condition, environment and situation. }\label{fig:tworobots}
\end{figure}

The Science of Resilient Robotic Autonomy deals with the research question of how to embed and encode robust, redundant and resourceful behaviors to autonomous robotic systems. This document is written at the moment subject to thoughts derived in the process of conducting research for robotics in extreme environments such as subterranean settings and nuclear facilities, agriculture services, as well as long-term off-road deployments. Figure~\ref{fig:tworobots} presents some of the environments that inspired the need to study further the questions discussed in this manuscript.

%
%

\section{A Closer Look to what enables Resilient Autonomy}

A Science of Resilient Robotic Autonomy, as envisioned in this document, may be approached by looking at each of its elements and then as a whole as it relates to any envisioned robotic functionality, namely a) resilient robotic embodiments, b) resilient navigation, c) resilience against internal system faults, d) resilience through the correlation of extrinsic and intrinsic objectives, e) resilience against environment threats and uncertainties, f) resilience against external attacks, g) resilience within and based on robot teams, as well as h) refining and enhancing resilience through incremental learning. This list is not exhaustive and not complete. It is however indicative and serves to outline concepts of thinking.

\subsection{Resilient Robotic Embodiments}\label{sec:embodiments}

The physical embodiment of a robot eventually dictates the limits of its capacities and functionality. Either we refer to a ground, aerial, surface, underwater or even space robotic system this was and remains a fact. Thus the science of resilient robotic autonomy has to start by a close look to the design principles required for resourcefulness, robustness and redundancy. With observations that relate to ubiquitous characteristics of natural organisms including the ability of animals to deal with varying terrain by adjusting their locomotion~\cite{alexander1990three}, resilience during physical interaction even on birds during their flight trajectories~\cite{pike2004scaling}, body flexibility and adaptability of most vertebrate species~\cite{dickinson2000animals}, as well as embedded sensing multi-modality across natural organisms~\cite{smith2008biology}, we could derive that some key elements of resilient robotic embodiments can relate to: a) robot mechanisms that afford or even exploit changes in the environment, b) robot mechanisms that afford unpredicted and unintentional physical interaction, c) robotic embodiments with diverse sensing skills and capacities. 

\subsection{Resilient Navigation}\label{sec:nav}

Seamless navigation in diverse environments is at the epicenter of robotic autonomy. Physical embodiment limitations-aside, autonomous navigation depends upon a) the robotic perception ability to enable robust localization and perform scene understanding~\cite{betge1994natural,durrant2006simultaneous,bailey2006simultaneous}, b) the onboard intelligence enabling motion planning and control for collision-free guidance~\cite{choset2005principles}, as well as c) the control robustness against diverse conditions~\cite{zhou1998essentials,slotine1991applied}. Resilient autonomous navigation thus may rely on a) resourceful localization, mapping and scene cognition through truly diverse sensing multi-modality, b) resourceful and robust motion planning cognizant to robot limitations and perception uncertainty/noise, as well as c) robust, redundant and fault-tolerant control. As these robot functional loops have to deal with diverse environments it is thought that they may benefit if designed accounting for their interconnection and couplings (``tightly closing the perception-action loop'') and if real-time or incremental (mission-to-mission) learning and meta-learning is facilitated. For the latter, identification of novelty and a mechanism for curiosity rewards would be essential.  

\subsection{Resilience against Internal System Faults}\label{sec:faultol}

Resilience depends on redundancy. Redundant actuation facilities fault-tolerant control~\cite{blanke2006diagnosis}. Redundant sensing and multi-modality is key for the potential of fusion algorithms that perform reliably not ``merely'' under every environment and subject to sensor stream degradation but also subject to sensor failure. Resilient autonomy requires a) tolerance against actuator failures, b) tolerance against sensor failures, c) tolerance against wear and extensive use in harsh conditions, d) tolerance versus developing drifts, biases, noise and other electronic propagation (especially in sensors), as well as e) knowledge of internal health states. The latter is particularly important. Invoking a policy to respond to an internal system fault requires observation of the respective states and detection-estimation of a failure and its condition. 

\subsection{Resilience through the correlation of Extrinsic and Intrinsic Objectives}

Robotic systems are assigned extrinsic objectives that relate to mission needs reflecting human interests. Exploration of unknown environments~\cite{dudek1978robotic}, infrastructure inspection~\cite{lattanzi2017review,sa2014vertical}, environmental monitoring~\cite{jadaliha2012environmental} and more are all indicative examples. In the process of optimizing for an extrinsic objective the robotic system decides the most appropriate paths and actions for that goal. However, most commonly when robots perform such operations they rarely relate their actions with monitoring of their overall internal health states. Although tracking of certain states such as battery level or actuator faults is there in the state-of-the-art, and despite major progress in belief-space planning~\cite{platt2010belief}, a generic and computationally tractable framework to co-optimize for both extrinsic objectives and intrinsic needs/sub-objectives is not available. Intrinsic objectives are however often critical and should not be neglected~\cite{oudeyer2016intrinsic}. The need to track and minimize estimation uncertainty during a mapping operation or navigation task~\cite{bry2011rapidly}, the importance of attentive perception~\cite{bajcsy2018revisiting}, or the value of team status observation during multi-robot teaming are all indicative examples. Resilience in robotic autonomy depends on the ability of the robot, through model-based approaches or learning to co-optimize for intrinsic objectives vital to its robust performance, without compromising - as much as physically possible - mission objectives.

\subsection{Resilience against Environment Threats and Uncertainties}\label{sec:env}

The more robotics become ubiquitous or undertake critical operations in challenging settings, the more their ability to predict the environment ahead will be decreased. Even as high-level reasoning improves, the complexity of certain environments and dynamic worlds can be overwhelming. For example, in the framework of autonomous driving~\cite{levinson2011towards} the world around the robot can be extremely dynamic and unpredictable especially as such systems will have to operate aside human drivers and pedestrians all around the world (dealing with different habits, urban planning principles etc). Resilient performance depends upon the ability of the robot to present robustness against unpredicted changes in the environment or threats to its navigation (e.g., dynamic obstacles). Multi-modality, powerful physical embodiments and advanced control/planning intelligence are all essential to achieve this goal. Going further, a key necessity is that robots should know when they cannot deal or even survive in a certain situation. Self-awareness of one’s limits is key to survival and reliable operation and should dictate how a mission needs be modified or even suspended. As the overall problem eventually relates to a complex Partially Observable Markov Decision Process, we owe to identify efficient but also generic methodologies to embed the ability for a robot to know and decide when continuing a mission is possible and if not how the task may change minimally to make it again feasible.

\subsection{Resilience against External Attacks}\label{sec:threatstl}

An issue less frequently encountered in the literature but in fact particularly important is the ability of a robot to present resilience against an external attack. Examples include but are not limited to a) the robustness of the perception system of an autonomous car against an adversarial attack to its sensor stream quality by otherwise ``subtle'' modifications of critical environmental components such as traffic signs~\cite{sitawarin2018rogue}, b) the ability of a house robot to present resilience in case of a maltreatment by its user, or c) the ability of a flying robot to survive an attack to its GPS data~\cite{kerns2014unmanned}. What type of desired resilient behavior should we want in such situations is not a simple question. Nevertheless, any resilient behavior starts from the ability of the robot to retain its capacity to estimate the state of the world and itself and some basic level of functionality. For example this is needed to ensure robust navigation even after a GPS attack if this the desired behavior or to know to halt its operation and wait in case of a house robotic agent, or to possibly refer to prior map knowledge of the traffic sign in the case of the aforementioned autonomous driving example. At least three types of resilience against external attacks are of significant interest, namely a) resilience against attacks that aim to take over or deteriorate digital control, b) resilience versus adversarial attacks that aim to confuse the inference of sensor streams, and c) some degree of resilience against physical damage to actuators. In deriving what are the desired policies of resilience in such conditions we must also refer to the philosophical discussion of how we would like to see robots widely utilized in our societies (the three laws of robotics from I. Asimov are always a good starting point to initiate a thinking process for that matter~\cite{murphy2009beyond,anderson2008asimov}). 

\subsection{Resilience in Robot Teams}\label{sec:teamedres}

Resilience is a property to be expressed and demonstrated both by individual robotic systems but also systems-of-systems, diverse robot teams of possibly heterogeneous agents. In fact, resilience is a characteristic that can be best achieved in the concept of robot teaming. First of all, this may relate to resilience benefits of multi-robot teams versus monolithic designs~\cite{parker1999adaptive}. In the case of an envisioned effective result one may be able to achieve the same both with a single complex robotic system or with a team of simpler systems. The latter inherently supports - in principle - redundancy in a more reliable fashion. Of particular importance is the question of how heterogeneity and diversity have a role and support the goal of facilitating resilience in robot teams. A second research goal is to address how robotic teamed resilience is established against attacks on one or more agents of the multi-robot system~\cite{saulnier2017resilient}. Third, another interesting challenge and goal relates to how to facilitate resilience (robustness, redundancy, and resourcefulness) through multi-robot distributed system design. How robotic ``mosaics'' can effect the desired result in multiple different ways and with many levels and layers of redundancy. Last but not least, the role of network security is intrinsic to the question of resilient autonomy in robotic teams. 

\subsection{Resilience through Incremental Learning}\label{sec:embodiments}

For us humans, learning is a continuous process. Our ability to resiliently handle a vast set of situations relies - in part - on our ability to learn through new experiences. If we assume that as robotic engineers we will face extreme difficulty in embedding a ``fixed'' intelligence and capabilities to handle every possible situation and environment, then incremental learning is what we may need to investigate deeply. Studies in curiosity and meta-learning~\cite{gottlieb2013information,oudeyer2004intelligent} form a solid scientific basis for such a challenging objective. This in turn will require novel contributions in the science of learning but also in the relevant hardware based on which experiences are captured (sensors) and processed (onboard or cloud compute).

\section{Towards a New Design Paradigm}

Eventually, the key question raised in this document relates to whether or not a new design paradigm for robotics is needed to achieve resilient autonomy. Answering this question precisely is quite hard and in fact this answer may only be found and discovered in and within the research to be done by the community. Nevertheless, there may be some observations worth making. 

The first relates to the value of (at least partially) departing from the traditional ``block-diagram approach'' of design within which every submodule is independently synthesized. Despite the extreme value of block-diagram based organization of functional loops, this is not a new discussion. Design of every functional loop cognizant to the limitations of the other is a recipe for enhanced resilience and it already partially takes place. To provide an example, a motion planning loop that accounts for the limitations of onboard localization will be able to provide the actions needed to support the observation process. In this framework, the theme of co-optimization of intrinsic and extrinsic objectives is - among others - particularly relevant. 

The second observation relates to the value of learning in robotics. Roboticists probably should still try to understand the systems and their environments as much as possible. Modeling and model-based design will retain its immense value. Yet in combination to the above there is a certain importance and a reasonable expectation for a new ``leap in resilience'' by learning. Behaviors that facilitate resilience may be difficult to be modeled, designed and predicted. The strength of data-driven approaches, especially when combined with model knowledge and model-based methods alongside novel research techniques and algorithms (e.g., in the foundations of reinforcement learning and unsupervised techniques) most likely will lead to a new rich set of important results. 

The third relates to highlighting the importance of novel robotic designs and the value of holistic design - conceptualizing robotic systems by thinking together their embodiment and their sensing and intelligence skills. New trends in domains such as legged locomotion, biomimetic flight or underwater navigation, soft robotics, event-driven vision, particularly sensitive multi-spectral imaging, neural computing and more are elements that when properly combined have a vast potential to advance robotic autonomy. As a community, we can benefit and accelerate robotics research if those dealing with every aspect of the above coordinate and collaborate. 

The above are only indicative of directions for a new design paradigm. The exact way, the new paradigm itself will be found by the research community (in its totality - even exceeding robotics) in the process. Yet there is one more important observation worth making in this discussion. Validating and verifying the progress will become an increasingly important aspect of the work as we deal with increasingly complex problems. In that sense, golden properties of the research will relate to a) reproducibility and b) replicability. The replication of scientific findings using independent investigators, methods, data, equipment, and protocols has long been, and will continue to be, the standard by which scientific claims are evaluated. Code releases and documented datasets are thus very important to support the process of systematic progress. In this framework, it would be of exceptional value to emphasize on field testing and evaluation or even in gathering and documenting objectively field experience as robotic systems are used in real-life situations. Such a procedure would facilitate the means needed to support the ``Science of Robotic Resilience'', and would guide us towards the future of ``Resilient and Field-hardened Robotic Autonomy''.

\section{Moving Forward}

This opinion manuscript serves to help in organizing the discussion for the themes, research topics and broadly ``what it takes'' to build a Science of Resilient Robotic Autonomy, a science of resilience to enable autonomous robots to navigate and operate seamlessly in diverse environments and settings thus enabling them to become ubiquitous within our societies. Iteratively revised versions of this manuscript will be released to better capture progress in the relevant domains or even the very basic thinking behind resilience and autonomy. 


\end{document}